\documentclass[12pt]{article}
\usepackage[margin=1in]{geometry}
\usepackage{amsmath,amssymb,amsthm}
\usepackage{graphicx}
\usepackage{booktabs}
\usepackage{hyperref}
\usepackage{cite}
\usepackage{algorithm}
\usepackage{algorithmic}
\usepackage{xcolor}
\usepackage{subcaption}

\newtheorem{theorem}{Theorem}

\title{\textbf{Deep Kuratowski Embedding Neural Networks for  Wasserstein Metric Learning}}

\author{Andrew Qing He\\
Department of Mathematics\\
Southern Methodist University\\
\texttt{andrewho@smu.edu}}

\date{\today}

\begin{document}

\maketitle

\begin{abstract}
Computing pairwise Wasserstein distances is a fundamental bottleneck in data analysis pipelines such as Wassmap~\cite{hamm2023wassmap} and LOT Wassmap~\cite{cloninger2025lot}. Motivated by the classical Kuratowski embedding theorem, we propose two neural architectures for learning to approximate the Wasserstein-2 distance ($W_2$) from data. The first, \emph{DeepKENN}, aggregates distances across all intermediate feature maps of a CNN using learnable positive weights. The second, \emph{ODE-KENN}, replaces the discrete layer stack with a Neural ODE, embedding each input into the infinite-dimensional Banach space $C^1([0,1], \mathbb{R}^d)$ and providing implicit regularization via trajectory smoothness. Experiments on MNIST with exact precomputed $W_2$ distances show that ODE-KENN achieves a 28\% lower test MSE than the single-layer baseline and 18\% lower than DeepKENN under matched parameter counts, while exhibiting a smaller generalization gap. The resulting fast surrogate can replace the expensive $W_2$ oracle in downstream pairwise distance computations.
\end{abstract}

\section{Introduction}
\label{sec:intro}

The Wasserstein distance, or Earth Mover's Distance, is a metric on the space of probability measures that has found widespread application in machine learning, image analysis, and data science~\cite{villani2008optimal,peyre2019computational}. Unlike divergences such as KL or total variation, the Wasserstein distance takes the underlying geometry of the sample space into account, making it well-suited for comparing structured objects such as images.

Despite its theoretical appeal, the practical computation of Wasserstein distances presents a significant bottleneck. Solving the exact linear program for $W_2$ between two discrete measures supported on $n$ points requires $O(n^3 \log n)$ time~\cite{bonneel2011displacement}, and even approximate methods such as Sinkhorn iteration~\cite{cuturi2013sinkhorn} remain expensive at scale.

This computational cost is particularly acute in algorithms that require the \emph{pairwise} $W_2$ distance matrix between a dataset of measures. A prominent example is Wassmap~\cite{hamm2023wassmap}, a nonlinear dimensionality reduction technique that embeds a dataset of images into a low-dimensional Euclidean space by applying Isomap to the pairwise $W_2$ distance matrix. Wassmap has been shown to exactly recover latent parameters of image manifolds generated by translations or dilations, but its scalability is limited by the cost of constructing the pairwise distance matrix. The subsequent LOT Wassmap~\cite{cloninger2025lot} addresses this by replacing exact $W_2$ with linearized optimal transport (LOT) approximations, providing approximation guarantees while avoiding the full pairwise matrix. Nevertheless, even LOT-based methods incur significant per-pair computational cost that could be alleviated by a fast learned surrogate.

\paragraph{Our contribution.} We propose and experimentally validate two neural architectures for learning the $W_2$ distance from a training set of measure pairs with precomputed ground-truth distances:

\begin{enumerate}
    \item \textbf{DeepKENN}: A convolutional neural network that computes distances as a weighted sum of squared Euclidean distances across all intermediate feature maps, with learnable positive weights enforced via a softplus reparameterization. This is motivated by the Kuratowski embedding theorem, which guarantees that any metric space embeds isometrically into a Banach space.

    \item \textbf{ODE-KENN}: An extension that replaces the discrete multi-layer sum with a Neural ODE on the feature space, producing a trajectory-integral distance. The embedding space is now infinite-dimensional ($C^1([0,1], \mathbb{R}^d)$), and the ODE's smoothness constraint acts as a natural regularizer.
\end{enumerate}

Both architectures are compared to a naive baseline (Euclidean distance in the final feature space) under matched parameter counts on MNIST $W_2$ distance learning.

\paragraph{Relationship to existing work.} Our multi-layer distance formulation is closely related to the perceptual loss of Johnson et al.~\cite{johnson2016perceptual}, which also aggregates feature-map differences across layers of a pretrained CNN for image synthesis tasks. Our contribution differs in that the weights are learned end-to-end alongside the backbone, and the goal is metric approximation rather than image generation. The Neural ODE component follows Chen et al.~\cite{chen2018neural}, with the key novelty that the \emph{distance} is defined via the trajectory integral rather than the terminal state.

\section{Background}
\label{sec:background}

\subsection{Wasserstein Distance}

Let $(\mathcal{X}, c)$ be a metric space and let $\mu, \nu \in \mathcal{P}_2(\mathcal{X})$ be probability measures with finite second moments. The Wasserstein-2 distance between $\mu$ and $\nu$ is
\begin{equation}
    W_2(\mu, \nu) = \left( \inf_{\gamma \in \Pi(\mu, \nu)} \int_{\mathcal{X} \times \mathcal{X}} c(x,y)^2 \, d\gamma(x,y) \right)^{1/2},
\end{equation}
where $\Pi(\mu, \nu)$ denotes the set of all couplings (joint distributions with marginals $\mu$ and $\nu$). For discrete measures supported on finite grids, this reduces to a linear program solvable by the network simplex method.

\subsection{Kuratowski Embedding}

The theoretical motivation for our approach is the following classical result.

\begin{theorem}[Kuratowski--Wojdys\l{}awski]
Every bounded metric space $(M, d)$ embeds isometrically as a closed subset of a convex set in the Banach space $\ell^\infty(M)$.
\end{theorem}

In practice, learning an isometric embedding into $\ell^\infty$ is computationally intractable due to its infinite dimensionality. The Johnson--Lindenstrauss lemma~\cite{johnson1984extensions} provides a finite-dimensional substitute: any $N$-point metric space embeds into $\mathbb{R}^k$ with distortion $1+\varepsilon$ for $k = O(\varepsilon^{-2} \log N)$. However, both approaches are passive — they describe \emph{existence} of embeddings, not how to learn one. Our work replaces the existence argument with a trainable embedding.

\subsection{Neural ODEs}

A Neural ODE~\cite{chen2018neural} defines the hidden state $h(t)$ of a network as the solution to the initial value problem
\begin{equation}
    \frac{dh}{dt} = \Phi_{\boldsymbol{\vartheta}}(h(t), t), \quad h(0) = h_0,
\end{equation}
where $\Phi_{\boldsymbol{\vartheta}}$ is a neural network. The output is $h(T)$ for some final time $T$. Neural ODEs can be viewed as the continuous-depth limit of ResNets, and gradients are computed efficiently via the adjoint method. A key property is that the map $h_0 \mapsto h(T)$ is a diffeomorphism when $\Phi_{\boldsymbol{\vartheta}}$ is Lipschitz, implying injectivity of the learned embedding.

\section{Deep Kuratowski Embedding Neural Network (DeepKENN)}
\label{sec:deepkenn}

\subsection{Architecture}

Let $F_{\boldsymbol{\vartheta}} : \mathcal{X} \to \mathbb{R}^{d_L}$ be a neural network with $L$ layers, so that
\[
    F_{\boldsymbol{\vartheta}} = f^{(L)}_{\boldsymbol{\vartheta}} \circ \cdots \circ f^{(1)}_{\boldsymbol{\vartheta}}.
\]
Denote the output of the $k$-th layer as $F^{(k)}_{\boldsymbol{\vartheta}}(x) := f^{(k)}_{\boldsymbol{\vartheta}} \circ \cdots \circ f^{(1)}_{\boldsymbol{\vartheta}}(x) \in \mathbb{R}^{d_k}$.

\paragraph{Naive baseline.} The simplest approach defines the distance as
\begin{equation}
    \hat{d}(x, \tilde{x}) = \left\| F^{(L)}_{\boldsymbol{\vartheta}}(x) - F^{(L)}_{\boldsymbol{\vartheta}}(\tilde{x}) \right\|_2.
    \label{eq:naive}
\end{equation}
This uses only the final layer representation of dimension $d_L$, discarding all intermediate structure.

\paragraph{DeepKENN.} We propose instead to use all intermediate representations:
\begin{equation}
    \hat{d}(x, \tilde{x}) = \sqrt{\sum_{k=1}^{L} \lambda_k \left\| F^{(k)}_{\boldsymbol{\vartheta}}(x) - F^{(k)}_{\boldsymbol{\vartheta}}(\tilde{x}) \right\|_2^2},
    \label{eq:deepkenn}
\end{equation}
where $\lambda_k \geq 0$ are learnable scalars. To enforce positivity, we parameterize $\lambda_k = \text{softplus}(\theta_k)$ for unconstrained $\theta_k \in \mathbb{R}$.

\paragraph{Embedding interpretation.} The distance~\eqref{eq:deepkenn} corresponds to a Euclidean distance in the product space $\mathbb{R}^{d_1} \times \cdots \times \mathbb{R}^{d_L}$, equipped with the weighted inner product $\langle u, v \rangle = \sum_k \lambda_k \langle u_k, v_k \rangle$. The effective embedding dimension is $d_{\text{tot}} = \sum_{k=1}^{L} d_k$, which can be substantially larger than $d_L$ alone. This is the discrete analog of the Kuratowski embedding.

\subsection{Triangle Inequality}

An important structural property is that~\eqref{eq:deepkenn} is guaranteed to satisfy the triangle inequality for any fixed backbone $F_{\boldsymbol{\vartheta}}$ and any $\lambda_k \geq 0$, since it is a Euclidean norm in the product embedding space. The positive definiteness (i.e., $\hat{d}(x,\tilde{x}) = 0 \Leftrightarrow x = \tilde{x}$) depends on the injectivity of the joint map $(F^{(1)}_{\boldsymbol{\vartheta}}, \ldots, F^{(L)}_{\boldsymbol{\vartheta}})$, which holds generically for sufficiently expressive networks.

\section{ODE-KENN}
\label{sec:odekenn}

\subsection{Motivation}

In DeepKENN, the effective embedding dimension $d_{\text{tot}} = \sum_k d_k$ is finite and grows only with the number of layers. The Kuratowski theorem, however, requires the embedding space to be infinite-dimensional for perfect isometry with an arbitrary metric space. This motivates replacing the discrete layer stack with a continuous-depth analog.

\subsection{Architecture}

ODE-KENN consists of two components: a CNN encoder $E_{\boldsymbol{\vartheta}} : \mathcal{X} \to \mathbb{R}^d$ that maps an input image to a fixed-dimensional feature vector, followed by a Neural ODE that evolves this feature vector continuously in time.

For input $x$, we define the trajectory $h_x : [0,1] \to \mathbb{R}^d$ as the solution to
\begin{equation}
    \frac{dh_x}{dt} = \Phi_{\boldsymbol{\vartheta}}(h_x(t)), \quad h_x(0) = E_{\boldsymbol{\vartheta}}(x),
    \label{eq:ode}
\end{equation}
where $\Phi_{\boldsymbol{\vartheta}} : \mathbb{R}^d \to \mathbb{R}^d$ is an autonomous MLP. The ODE-KENN distance is then defined as
\begin{equation}
    \hat{d}(x, \tilde{x}) = \sqrt{\int_0^1 \lambda_t \left\| h_x(t) - h_{\tilde{x}}(t) \right\|_2^2 \, dt},
    \label{eq:odekenn}
\end{equation}
where $\lambda_t \geq 0$ is a learnable weight function discretized over the ODE solver's time steps, again parameterized via softplus.

\subsection{Infinite-Dimensional Embedding}

Each input $x$ is mapped to a curve $h_x \in C^1([0,1], \mathbb{R}^d)$. The Banach space $C^1([0,1], \mathbb{R}^d)$ is infinite-dimensional, and the map $x \mapsto h_x$ is injective when $E_{\boldsymbol{\vartheta}}$ is injective (since distinct initial conditions $h_x(0) \neq h_{\tilde{x}}(0)$ produce distinct trajectories under a Lipschitz vector field). The distance~\eqref{eq:odekenn} then measures the weighted $L^2$ separation of these trajectories.

\subsection{Implicit Regularization}

The Lipschitz regularity of $\Phi_{\boldsymbol{\vartheta}}$ controls the smoothness of trajectories. Specifically, if $\Phi_{\boldsymbol{\vartheta}}$ is $L$-Lipschitz, then
\[
    \left\| h_x(t) - h_{\tilde{x}}(t) \right\|_2 \leq e^{Lt} \left\| h_x(0) - h_{\tilde{x}}(0) \right\|_2,
\]
which bounds the trajectory separation in terms of the initial feature difference. This exponential bound acts as a regularizer: it prevents the ODE from producing arbitrarily large intermediate differences that could overfit to training pairs.

\subsection{Practical Implementation}

We solve the ODE using the RK4 method with fixed step size via \texttt{torchdiffeq}~\cite{chen2018neural}. To avoid two separate ODE solves per training pair, we stack the two initial conditions on the batch dimension:
\[
    h_0^{\text{joint}} = [h_x(0); h_{\tilde{x}}(0)] \in \mathbb{R}^{2B \times d},
\]
and solve a single ODE of doubled batch size. The trajectory is then split and the distance computed. This halves the ODE evaluation cost per pair.

\section{Experimental Setup}
\label{sec:experiments}

\subsection{Dataset}

We use the MNIST handwritten digit dataset~\cite{lecun1998mnist}, treating each $28 \times 28$ grayscale image as a discrete probability measure by normalizing pixel intensities to sum to one. The ground cost matrix is $C_{ij} = \|p_i - p_j\|_2^2$ where $p_i, p_j \in \mathbb{R}^2$ are pixel coordinates, so neighboring pixels have unit squared distance.

We precompute exact $W_2$ distances for 55,000 image pairs using \texttt{ot.emd2} from the Python Optimal Transport library~\cite{flamary2021pot}, covering 1,000 random pairs for each of the $\binom{10}{2} + 10 = 55$ digit class combinations $(i, j)$ with $0 \leq i \leq j \leq 9$. The resulting dataset is split into 49,500 training, 2,750 validation, and 2,750 test pairs.

\subsection{Network Architecture}

All three models share an identical CNN encoder:
\[
\begin{aligned}
    &\text{Conv}(1 \to 8, 5\times5) \to \text{Pool} \to \text{Conv}(8 \to 16, 3\times3) \to \text{Pool} \\ &\to \text{Conv}(16 \to 32, 3\times3) \to \text{Pool} \to \text{FC}(288 \to 128) \to \text{FC}(128 \to 64).
\end{aligned}
\]
The encoder outputs a 64-dimensional feature vector. The intermediate feature dimensions are $d_1 = 1568$, $d_2 = 784$, $d_3 = 288$, $d_4 = 128$, $d_5 = 64$, giving $d_{\text{tot}} = 2832$ for DeepKENN.

To ensure a fair comparison with matched parameter counts, both Naive and DeepKENN are augmented with an additional \texttt{FC}$(64 \to 64)$+\texttt{Tanh} head (4,160 parameters), matching the ODE function's parameter count. Total parameter counts: Naive 55,424; DeepKENN 55,430; ODE-KENN 55,434. The ODE operates with $N = 10$ time steps over $[0,1]$ using RK4.

\subsection{Training}

All models are trained with the Adam optimizer, learning rate $10^{-3}$, batch size 256, for 2,000 epochs, using the MSE loss between predicted and true $W_2$ distances. The best validation checkpoint is used for final evaluation.

\section{Results}
\label{sec:results}

\subsection{Quantitative Performance}

\begin{table}[h]
\centering
\caption{Test set performance on MNIST $W_2$ distance learning (2,750 pairs). Distances are in pixel units. Rel.\ MAE = MAE / mean$(W_2)$.}
\label{tab:results}
\begin{tabular}{lccc}
\toprule
Model & Test MSE & Test MAE & Rel.\ MAE \\
\midrule
Naive (single-layer) & $4.67 \times 10^{-3}$ & $5.249 \times 10^{-2}$ & 1.63\% \\
DeepKENN & $4.07 \times 10^{-3}$ & $4.873 \times 10^{-2}$ & 1.52\% \\
\textbf{ODE-KENN} & $\mathbf{3.35 \times 10^{-3}}$ & $\mathbf{4.406 \times 10^{-2}}$ & \textbf{1.37\%} \\
\bottomrule
\end{tabular}
\end{table}

Table~\ref{tab:results} summarizes the test performance. ODE-KENN achieves a 28\% reduction in MSE relative to the Naive baseline and an 18\% reduction relative to DeepKENN, despite identical parameter budgets.

\begin{figure}[!htbp]
    \centering
    \includegraphics[width=\textwidth]{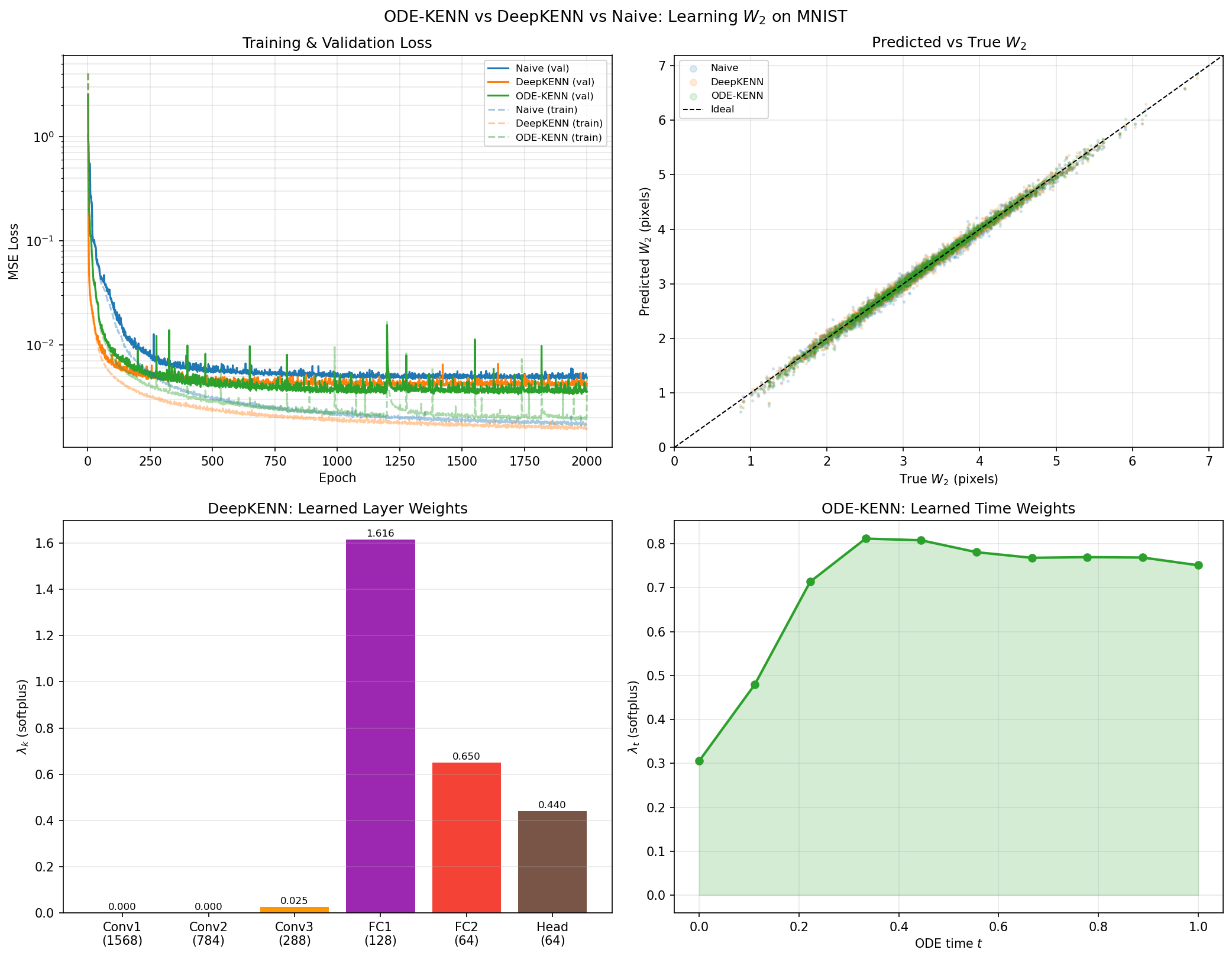}
    \caption{Experimental results for all three models trained for 2,000 epochs on MNIST $W_2$ distance learning.
    \textbf{Top left:} Training (dashed) and validation (solid) MSE loss curves on a log scale. ODE-KENN (green) converges fastest and maintains the smallest train/val gap throughout training.
    \textbf{Top right:} Predicted vs.\ true $W_2$ distances on the test set (2,750 pairs). ODE-KENN produces the tightest scatter around the ideal diagonal, particularly at small $W_2$ values.
    \textbf{Bottom left:} Learned layer weights $\lambda_k$ for DeepKENN. The first two convolutional layers are nearly fully suppressed ($\lambda \approx 0$), while FC1 dominates ($\lambda \approx 1.62$), indicating that abstract compressed representations are most informative for $W_2$.
    \textbf{Bottom right:} Learned time weights $\lambda_t$ for ODE-KENN over $t \in [0,1]$. The bell-shaped profile peaks near $t \approx 0.35$, downweighting the raw feature ($t=0$) and the late trajectory, and assigning greatest importance to the early-to-mid ODE evolution.}
    \label{fig:results}
\end{figure}

\subsection{Convergence and Overfitting}
\label{sec:convergence}

Figure~\ref{fig:results} (top left) shows the training and validation loss curves for all three models. All models converge by epoch 2,000, but with notably different train/val gaps. ODE-KENN exhibits the smallest generalization gap among the three models, consistent with the implicit regularization provided by the ODE's trajectory smoothness. DeepKENN shows a larger gap, reflecting its higher effective embedding dimension ($d_{\text{tot}} = 2832$) compared to the single-layer baseline ($d = 64$) and ODE-KENN's smooth trajectory structure.

\subsection{Scatter Plot Analysis}

Figure~\ref{fig:results} (top right) shows predicted versus true $W_2$ values on the test set. All models track the diagonal well across the full range of $W_2$ values (approximately 0.5 to 6.5 pixels). ODE-KENN produces the tightest scatter, particularly at small $W_2$ values (below 1.5 pixels), which correspond to within-class pairs and are the most geometrically important for downstream dimensionality reduction.

\subsection{Learned Weight Analysis}

\paragraph{DeepKENN layer weights.} Figure~\ref{fig:results} (bottom left) shows the learned $\lambda_k$ values for DeepKENN. The weights decay dramatically across the convolutional layers: Conv1 ($\lambda \approx 0$), Conv2 ($\lambda \approx 0$), Conv3 ($\lambda \approx 0.025$), while the fully connected layers dominate with FC1 ($\lambda \approx 1.62$) and FC2 ($\lambda \approx 0.65$). This reveals that, for $W_2$ on MNIST images, pixel-scale spatial features are nearly uninformative; the abstract compressed representations from the FC layers carry essentially all the metric information.

\paragraph{ODE-KENN time weights.} The learned $\lambda_t$ profile for ODE-KENN (Figure~\ref{fig:results}, bottom right) rises steeply from approximately 0.31 at $t=0$ to a peak near $t \approx 0.35$, then gradually decreases to about 0.75 at $t=1$. This bell-shaped profile indicates that the network assigns greatest weight to the early-to-middle portion of the trajectory. The initial condition ($t=0$, raw CNN feature) is downweighted, while the ODE-transformed representation at intermediate times carries the most discriminative information for $W_2$ approximation.

\section{Application to Wasserstein Dimensionality Reduction}
\label{sec:application}

A primary motivation for this work is to accelerate pipelines that require pairwise $W_2$ distance matrices. We briefly describe how our framework integrates with existing methods.

\paragraph{Wassmap.} Wassmap~\cite{hamm2023wassmap} applies Isomap to a pairwise $W_2$ distance matrix $\mathbf{D}$ of $N$ images, where $D_{ij} = W_2(\mu_i, \mu_j)$. For $N = 1000$ images on a $28 \times 28$ grid, constructing $\mathbf{D}$ exactly requires $\binom{1000}{2} = 499{,}500$ calls to the linear program solver, each taking approximately 5 milliseconds — totaling roughly 42 minutes on a single CPU core. With a trained ODE-KENN model, each pairwise distance evaluation requires a single forward pass, reducing the evaluation to seconds. The approximation error (Rel.\ MAE $\approx 1.37\%$) introduces a bounded perturbation to the distance matrix that can be bounded in terms of the MDS sensitivity analysis of~\cite{cloninger2025lot}.

\paragraph{LOT Wassmap.} LOT Wassmap~\cite{cloninger2025lot} avoids the full pairwise matrix by working with linearized transport maps. Our method is complementary: while LOT Wassmap provides theoretical guarantees on embedding quality under approximation, ODE-KENN provides a data-driven alternative that may better exploit the structure of specific image families (e.g., handwritten digits) at the cost of requiring a training set.

\paragraph{Generalization.} Our trained model generalizes to new image pairs at inference time without retraining, making it suitable as a drop-in replacement for the $W_2$ oracle in any algorithm requiring pairwise distances. The training cost — approximately one hour for 2,000 epochs on a single GPU — is a one-time investment amortized over all subsequent evaluations.

\section{Discussion}
\label{sec:discussion}

\paragraph{Metric structure.} The distance formulas~\eqref{eq:deepkenn} and~\eqref{eq:odekenn} are guaranteed to satisfy the triangle inequality and non-negativity by construction, since they are (weighted) $L^2$ norms of embedding differences. Positive definiteness depends on the injectivity of the embedding, which holds generically for trained networks but is not guaranteed.

\paragraph{The role of the CNN encoder.} Our experiments reveal that the CNN encoder's convolutional layers contribute almost nothing to the DeepKENN distance ($\lambda_1 \approx \lambda_2 \approx 0$). This strongly suggests that the appropriate architecture for $W_2$ learning is a deep encoder that produces a compressed semantic representation, followed by a metric-learning module (either the DeepKENN sum or the ODE trajectory). The spatial hierarchy of convolutional features appears unnecessary for this task.

\paragraph{ODE regularity and generalization.} The gap between ODE-KENN's train and validation loss is smaller than for DeepKENN throughout training. We attribute this to the smoothness constraint on the ODE vector field: because $h_x(t)$ must follow a differentiable path, the model cannot memorize arbitrary per-sample distances through large trajectory oscillations. This is analogous to weight norm regularization in classical networks, but operates on the function space of trajectories rather than parameter space.

\paragraph{Limitations.} The current approach requires a precomputed training set of exact $W_2$ distances, which is expensive to produce (approximately 4–5 minutes for 55,000 pairs with images preloaded in RAM). For new image domains, this cost must be paid upfront. Additionally, the model is not guaranteed to produce a true metric (positive definiteness may fail at test time), which could affect downstream algorithms that rely on metric properties.

\section{Conclusion}
\label{sec:conclusion}

We have proposed DeepKENN and ODE-KENN, two neural network architectures for learning to approximate the $W_2$ Wasserstein distance from data. Both architectures are motivated by the Kuratowski embedding theorem and offer progressively richer embedding spaces: DeepKENN embeds into a finite-dimensional product space of dimension $d_{\text{tot}} = \sum_k d_k$, while ODE-KENN embeds into the infinite-dimensional Banach space $C^1([0,1], \mathbb{R}^d)$.

Experiments on MNIST demonstrate that ODE-KENN outperforms both DeepKENN and the naive baseline under matched parameter counts, achieving a 28\% reduction in test MSE and exhibiting better generalization throughout training. The learned weight profiles confirm our theoretical intuition: ODE-KENN downweights the raw feature representation and upweights intermediate trajectory states, while DeepKENN learns to suppress early convolutional features in favor of abstract FC representations.

Our framework provides a practical tool for accelerating pairwise $W_2$ computations in downstream applications such as Wassmap~\cite{hamm2023wassmap} and LOT Wassmap~\cite{cloninger2025lot}, reducing per-pair evaluation from milliseconds (exact LP) to microseconds (neural forward pass) after a one-time training investment.

Future work includes: (1) extending to $W_1$ and other $p$-Wasserstein distances; (2) theoretical analysis of approximation error in terms of network capacity and training set size; (3) application to higher-dimensional images beyond MNIST; and (4) incorporating the metric properties (triangle inequality, positive definiteness) as explicit training constraints.

\section*{Acknowledgments}

The author thanks Professor Keaton Hamm (University of Texas at Arlington) for inspiring discussions on Wasserstein-based dimensionality reduction and the computational challenges of pairwise distance matrix construction.

\bibliographystyle{plain}
\bibliography{references}

\appendix

\section{Architecture Details}
\label{app:arch}

\begin{table}[h]
\centering
\caption{CNN encoder architecture shared by all three models.}
\begin{tabular}{llll}
\toprule
Layer & Operation & Output shape & Flat dim \\
\midrule
Conv1 & Conv$(1\to8, 5\times5)$ + ReLU + MaxPool$(2)$ & $(8, 14, 14)$ & 1568 \\
Conv2 & Conv$(8\to16, 3\times3)$ + ReLU + MaxPool$(2)$ & $(16, 7, 7)$ & 784 \\
Conv3 & Conv$(16\to32, 3\times3)$ + ReLU + MaxPool$(2)$ & $(32, 3, 3)$ & 288 \\
FC1 & Linear$(288\to128)$ + ReLU & $(128,)$ & 128 \\
FC2 & Linear$(128\to64)$ & $(64,)$ & 64 \\
\bottomrule
\end{tabular}
\end{table}

\noindent ODE-KENN vector field: $\Phi_{\boldsymbol{\vartheta}}(h) = \tanh(Wh + b)$, with $W \in \mathbb{R}^{64\times64}$, $b \in \mathbb{R}^{64}$ (4,160 parameters). The Naive and DeepKENN models include an identical \texttt{FC}$(64\to64)$+\texttt{Tanh} head to match this count.

\end{document}